\documentclass[11pt]{article}

\usepackage[preprint]{acl}

\usepackage{times}
\usepackage{latexsym}
\usepackage[utf8]{inputenc}
\usepackage[T1]{fontenc}
\usepackage{booktabs}
\usepackage{amsfonts}
\usepackage{nicefrac}
\usepackage{microtype}
\usepackage{inconsolata}
\usepackage{xcolor}
\usepackage{listings}
\usepackage{graphicx}

\usepackage{multirow}
\usepackage{subfigure}
\usepackage[english]{babel}
\usepackage{moresize}
\usepackage{amsmath}
\usepackage{algorithmic}
\usepackage{balance}
\usepackage{comment}
\usepackage{paralist}
\usepackage{bm}
\usepackage{pgfplots}
\usetikzlibrary{pgfplots.dateplot}
\usepackage{flushend}

\usepackage{amssymb}
\usepackage{bbm}
\usepackage{longtable}
\usepackage{rotating}
\usepackage{mathrsfs}
\usepackage{enumitem}
\usepackage[linesnumbered,algoruled,boxed,lined]{algorithm2e}
\usepackage{adjustbox}
\usepackage{filecontents}
\usepackage{makecell}

\lstdefinestyle{promptstyle}{
  basicstyle=\ttfamily\footnotesize,
  breaklines=true,
  breakatwhitespace=true,
  columns=fullflexible,
  keepspaces=true,
  showstringspaces=false,
  frame=single,
  captionpos=b,
  lineskip=1pt,
  aboveskip=6pt,
  belowskip=6pt
}

\def\model{ViMax}

\title{ViMax: Agentic Video Generation}

\author{
  Lingxuan Huang$^{1*}$, Sizhe He$^{2*}$, Hengji Zhou$^{2*}$, Liqiang Nie$^3$,  Lianghao Xia$^{3\dagger}$, Chao Huang$^{1\dagger}$ \\
  $^1$The University of Hong Kong, $^2$South China University of Technology, \\
  $^3$Harbin Institute of Technology, Shenzhen \\
  \texttt{xvrhuang@connect.hku.hk}, \texttt{202320143418@mail.scut.edu.cn}, \\
  \texttt{hengjizhou01@gmail.com}, \texttt{xialh@hit.edu.cn}, \texttt{chaohuang75@gmail.com}
}

\begin{document}

\maketitle
\footnotetext[1]{$^*$Lingxuan Huang, Sizhe He, and Hengji Zhou have equal contribution to this work.}
\footnotetext[2]{$^\dagger$Lianghao Xia and Chao Huang are the corresponding authors.}

\begin{abstract}
Long-form video generation requires systematic narrative planning and visual consistency that current short-clip methods cannot provide. Existing methods generate isolated sequences without narrative structure and lack mechanisms for maintaining character and environmental consistency across scenes. We present \model\, an agentic video generation framework that addresses video creation through coordinated multi-agent collaboration where specialized components negotiate narrative decisions, visual continuity, and production quality. Our framework employs a hierarchical narrative engine with retrieval-augmented generation for global story coherence and a dependency-aware visual consistency mechanism that tracks character and environmental states across temporal boundaries, while VLM-guided agents continuously monitor and refine both narrative coherence and visual fidelity. The framework enables coordinated agent collaboration to generate extended narrative content. This maintains both storytelling integrity and visual coherence across multi-scene timelines. The code for \model\ is available at: \href{https://github.com/HKUDS/ViMax}{https://github.com/HKUDS/ViMax}.
\end{abstract}

\section{Introduction} 
\label{sec:intro} 
Current video generation methods face critical limitations, producing only short clips without narrative structure while character appearances and environmental elements change unpredictably across scenes. Recent breakthroughs in generative AI offer transformative potential to address these fundamental challenges by automating complex video production pipelines through intelligent agentic frameworks, enabling creators to transform conceptual ideas into coherent long-form visual narratives with unprecedented ease.
\begin{figure*}[t]
  \centering
  \includegraphics[width=\textwidth]{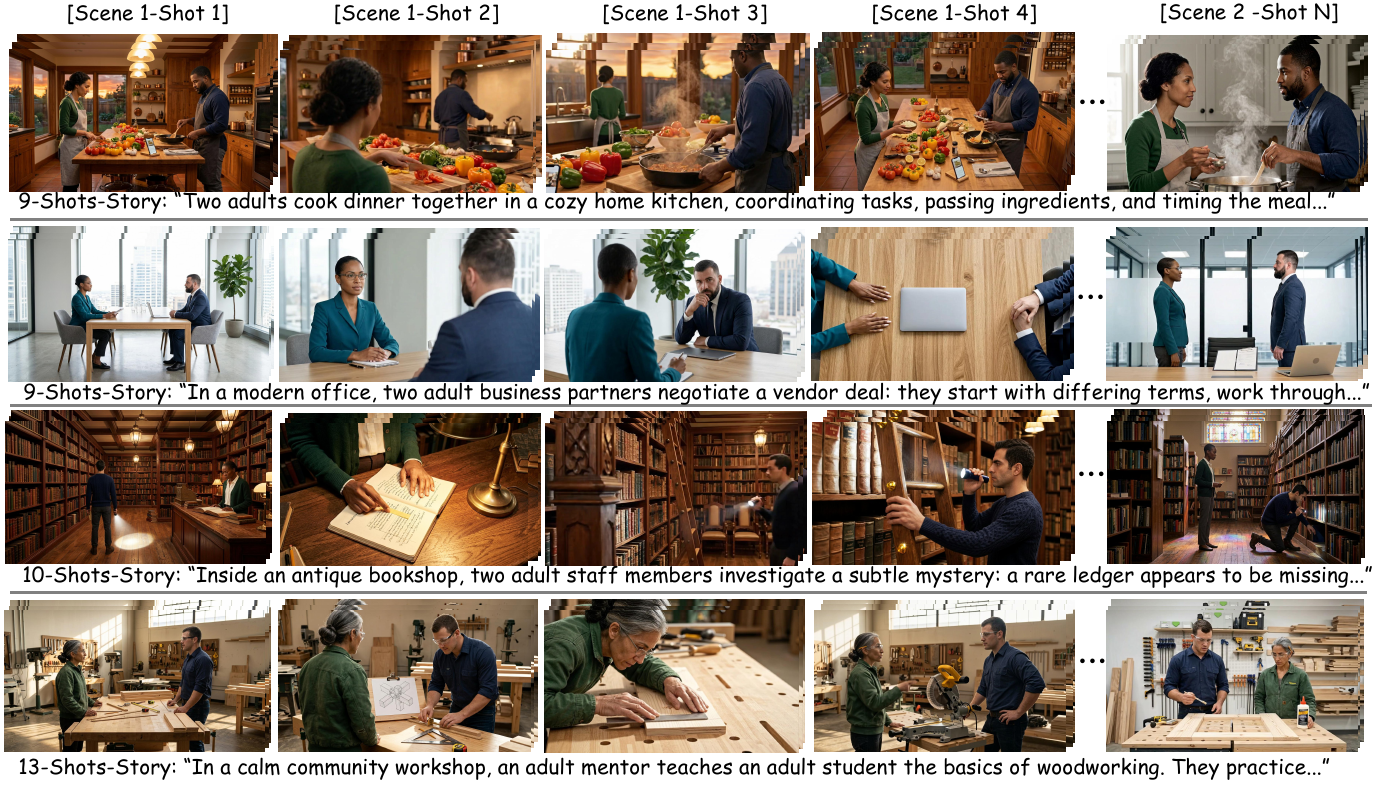}
  \vspace{-0.14in}
  \caption{Multi-shot video examples generated by \model\ on ViMax-Bench.}
  \label{fig:vimax_examples}
  \vspace{-0.06in}
\end{figure*}

Recent research in video generation has explored multiple directions to address these challenges. Text-to-video models such as Sora~\cite{OpenAI2024Sora}, Veo 3~\cite{deepmind2024veo}, and Seedance~\cite{Seedance2026Seedance2A}, together with open-source efforts like HunyuanVideo~\cite{kong2024hunyuanvideo} and Wan~\cite{Wang2025WanOA}, have achieved impressive per-clip fidelity. However, these models remain constrained to short-form content lasting only seconds. When extended to longer horizons, they suffer from \emph{catastrophic semantic forgetting}, where character identity, world geometry, and physical laws drift unpredictably across shots~\cite{kupyn2025epipolar,rakheja2025world}, breaking narrative immersion.

Two approaches have emerged to tackle temporal consistency, yet both face significant limitations. Streaming methods (StreamingT2V~\cite{Henschel2025StreamingT2V}, FreeLong~\cite{Lu2024FreeLong}) extend single-clip duration but cannot handle multi-scene narratives. Consistency mechanisms such as StoryDiffusion~\cite{zhou2024storydiffusion}, ConsiStory~\cite{tewel2024consistory}, and STAGE~\cite{Zhang2025STAGESG} preserve subject identity through shared attention or keyframe anchoring but lack narrative planning. Multi-agent frameworks including FilmAgent~\cite{xu2025filmagent}, Kubrick~\cite{He2024Kubrick}, and VideoGen-of-Thought~\cite{zheng2025videogen} decompose production into specialized roles but fail to coordinate complex narrative, visual, and temporal constraints. Existing systems struggle to maintain global coherence across dozens of shots required for long-form storytelling.

Addressing these limitations requires overcoming two fundamental challenges:\vspace{-0.1in}
\begin{itemize}[leftmargin=*]

\item \textbf{Multi-shot Visual Consistency.} Current video generators operate on isolated shots without awareness of previous content, causing character appearance, environment design, and spatial layout to drift. Long-form production demands object-level consistency to preserve character and environment properties, and spatial coherence to maintain consistent 3D geometry across angles.


\item \textbf{Long-form Narrative Planning.} Orchestrating narratives spanning dozens of shots exceeds current language models' reasoning capacity when treated monolithically. The system needs to coordinate character arcs, plot development, and thematic coherence while keeping decisions grounded in broader story context.

\end{itemize}

To address these challenges, we present \model, an agentic framework that orchestrates long-form video production from conceptual ideas to fully rendered content. \model\ decomposes this pipeline into specialized agents for screenwriting, shot planning, character styling, video generation, and VLM-based quality control with best-of-$k$ selection.

For narrative complexity, \model\ employs hierarchical story decomposition with retrieval-augmented generation. This keeps local planning grounded in global context. For visual consistency, \model\ introduces a graph-based dependency system that tracks cross-shot relationships during planning. Transition video generation maintains spatial coherence across camera angles within scenes.



Our primary contributions in this work include:\vspace{-0.05in}
\begin{itemize}[leftmargin=*]
\item \textbf{Agentic Long-form Video Generation Framework.} We introduce a multi-agent architecture that coordinates screenwriting, shot planning, character styling, video generation, and VLM-based quality control. This framework automates narrative video creation from conceptual ideas.

\item \textbf{Hierarchical Narrative Planning with Global Context.} We propose hierarchical story decomposition paired with retrieval-augmented generation. This enables language models to maintain thematic coherence and long-range dependencies across extended narratives.

\item \textbf{Graph-based Visual Consistency.} We address visual consistency through dual-level mechanisms. Graph-based dependency tracking ensures object consistency, while transition video generation maintains spatial coherence.
\end{itemize}

\section{Methodology}
\label{sec:solution} 
\begin{figure*}[t]
  \centering
  \includegraphics[width=\linewidth]{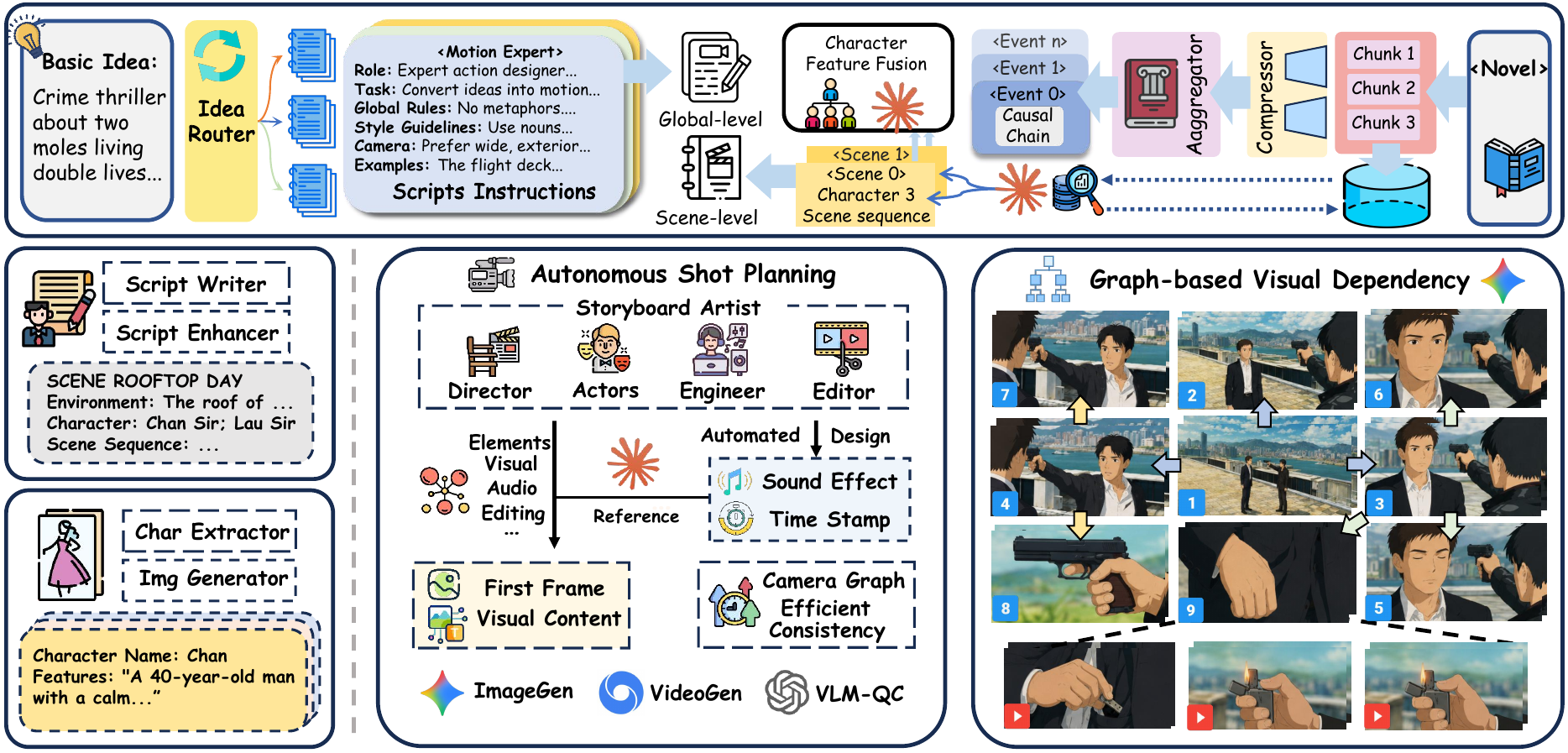}
  \vspace{-0.12in}
  \caption{The \model\ agentic video generation architecture.} 
  \label{fig:framework}
  \vspace{-0.08in}
\end{figure*}

This section elaborates technical details of \model. The overall architecture is depicted in Figure~\ref{fig:framework}.

\subsection{Long-Form Video Generation Workflow}
To generate long-form videos, \model\ employs a sophisticated multi-agent workflow that addresses the inherent complexity of extended narrative content creation. Unlike short-form video generation, long-form production demands coordination across multiple aspects including narrative structure, character consistency, cinematographic design, and quality control. Our workflow encompasses key production roles including directorial vision, screenwriting, shot planning, character styling, video generation, and quality assessment through specialized agents that collaborate to maintain coherence across the entire video production pipeline.

\subsubsection{From Ideas to Videos}
\noindent\textbf{Screenwriting}.
Users first state their creation ideas using natural language. This input $\mathcal{I}$ encompasses various narrative forms, including high-level creative intentions described in a few sentences such as "a detective solving a mystery in a small town", complete novels, or existing screenplay drafts.

To determine the video's storyline, \model\ first projects the user input $\mathcal{I}$ into a screenplay with unified format. This process can be described as:
\begin{align}
    \mathcal{S} = \text{ScreenWriting}(\mathcal{I}; P_{C(\mathcal{I})})
\end{align}
where $\text{ScreenWriting}(\cdot)$ is the LLM-based screenwriting method that composes high-quality scripts based on user input $\mathcal{I}$. This process is driven by different prompts $P_{C(\mathcal{I})}$ tailored to different categories of user input, and a categorization mechanism $C(\cdot)$ routes the user input into appropriate video creation categories. The resulting screenplay $\mathcal{S}$ provides a structured foundation with scene descriptions, character dialogues, and narrative progression that guides subsequent production stages.
\vspace{0.05in}

\noindent\textbf{Shot Planning}. To determine the appropriate cinematic language and visual expression for conveying the story's narrative, \model\ then conducts shot planning with our storyboard agent. It is powered by carefully crafted prompts that encapsulate professional cinematographic expertise and shot design principles. The agent provides detailed descriptions of visual content, sound effects, and dialogue for each shot in a structured sequence. Formally, this process can be described as follows:
\begin{align}
    S_1, S_2,\cdots,S_n = \text{ShotPlanning}(\mathcal{S})
\end{align}
where $S_1, S_2, \cdots, S_n$ represents the sequence of shots planned to complete the video's narrative, and $\text{ShotPlanning}(\cdot)$ is an LLM-based function that transforms the screenplay $\mathcal{S}$ into detailed shot specifications. Each shot $S_i$ contains comprehensive production details including camera angles, movements, lighting conditions, character positioning, and temporal duration, providing the technical blueprint for subsequent video generation stages.

\vspace{0.05in}
\noindent\textbf{Text-Image-Video Generation}. To better control the generation process, \model\ adopts a two-step text-image-video generation process. It first generates key images such as character profiles, scene backgrounds, and key frames of shots. This process includes text-to-image generation and text-and-image-to-image generation, as some images such as shot key frames require other images such as characters as reference. The second process utilizes text-and-image-to-video generation tools to generate videos for each shot. This process can be characterized as follows:
\begin{align}
    &\{\mathcal{K}_1, \mathcal{K}_2, \cdots, \mathcal{K}_m\} = \text{ImageGen}(\{S\}) \nonumber\\
    &\{V_1, V_2, \cdots, V_n\} = \text{VideoGen}(\{S\}, \{\mathcal{K}\})
\end{align}
where $\{\mathcal{K}_1, \mathcal{K}_2, \cdots, \mathcal{K}_m\}$ represents the set of generated key images including character designs and scene references, and $\{V_1, V_2, \cdots, V_n\}$ represents the final video segments corresponding to each planned shot. This process enhances \model's control over the final generated videos through explicit specification of key image content and quality control at the intermediate visual stage.

\subsubsection{VLM-based Quality Control}
To ensure generation quality, \model\ employs a VLM-based assessment mechanism that performs \emph{best-of-$k$ candidate selection}. For each task, the system samples $k$ candidates in parallel and uses a VLM judge to score them against the shot planning requirements along dimensions including visual fidelity, narrative consistency, and shot specification adherence. The highest-scoring candidate then advances through the pipeline. This process is formalized as:
\begin{align}
    \{C_1, C_2, \cdots, C_k\} &= \text{MultiGen}(S_i, \text{prompt}) \nonumber\\
    \text{score}_j &= \text{VLM-Judge}(C_j, S_i) \nonumber\\
    C^* &= \arg\max_{C_j} \text{score}_j 
\end{align}
where $\{C_1, \cdots, C_k\}$ are the $k$ sampled candidates, $\text{VLM-Judge}(\cdot)$ scores each against the shot requirements $S_i$, and $C^*$ is the selected best candidate. This selection mitigates the variance of single-shot generation and ensures that only the most faithful output progresses through the long-form pipeline.

\subsection{Long-Form Narrative Planning}
Long-form narrative planning presents a fundamental challenge due to the exponential complexity of coordinating multiple story elements across extended timelines. Unlike short-form content, long-form videos require jointly orchestrating character arcs, plot development, and narrative coherence spanning hundreds of scenes. The scale of information substantially exceeds current language models' reasoning capacity when approached as a single planning task, while maintaining global consistency across localized scenes creates tension between detailed control and overarching coherence.

\subsubsection{Hierarchical Story Decomposition}
To address the structural complexity inherent in long-form video planning, \model\ employs a hierarchical story decomposition approach. Long-form videos involve hundreds and even thousands of shots that exhibit clear hierarchical structure: multiple shots may collectively represent a single action sequence or dialogue exchange, while shots occurring within the same space share strong visual dependencies for lighting, character positioning, and environmental consistency. This nested structure makes direct shot-level planning computationally intractable and prone to inconsistencies.

Our hierarchical decomposition systematically breaks down the complete storyline into manageable planning units through recursive decomposition. Starting from the complete screenplay, the system first identifies major story events that represent significant narrative milestones. Each event is then decomposed into constituent scenes that capture specific dramatic moments or location changes. Finally, individual scenes are broken down into detailed shot specifications that define camera angles, character actions, and visual elements. Formally, this recursive process can be expressed as:
\begin{align}
{N_{d,1}, N_{d,2}, \cdots, N_{d,k_d}} = \text{Decomp}(N_{d-1,i})
\end{align}
where $N_{d-1,i}$ represents a parent node at depth $d-1$, ${N_{d,1}, N_{d,2}, \cdots, N_{d,k_d}}$ represents its child nodes at depth $d$. Each node $N_{d,j}$ is associated with a textual narrative description $T_{d,j}$. This approach ensures that language models tackle appropriately scoped planning tasks at each level, avoiding the cognitive overload of simultaneous multi-scale narrative design while preserving the natural hierarchical dependencies between story elements.

\subsubsection{RAG-based Global Context Awareness}
To address the information loss inherent in hierarchical decomposition, \model\ employs a RAG-enhanced approach that maintains global context awareness throughout the planning process. While hierarchical decomposition effectively manages complexity, it introduces a critical limitation: the generated descriptions for individual nodes primarily focus on defining the scope and boundaries of each narrative unit rather than providing rich, detailed content descriptions. Consequently, valuable contextual information from the original source material, such as character motivations, subtle plot connections, and thematic elements, is often discarded during the decomposition process.

Our solution integrates a retrieval-augmented generation (RAG) system that preserves access to the complete source material throughout the planning pipeline. The system first creates a comprehensive index of the original screenplay or novel content, then dynamically retrieves relevant context based on the textual descriptions generated during decomposition. For each narrative node, the system queries the indexed source material to enrich local planning with pertinent global information. Formally, this process can be described as:
\begin{align}
    \mathcal{R} = \text{Index}&(\mathcal{S}),~~
    C_{d,j} = \text{Retrieve}(\mathcal{R}, T_{d,j})\nonumber\\
    &T'_{d,j} = \text{Enrich}(T_{d,j}, C_{d,j})
\end{align}
where $\mathcal{R}$ represents the indexed representation of the source material, $C_{d,j}$ denotes the retrieved contextual information relevant to node description $T_{d,j}$, and $T'_{d,j}$ is the enriched description that incorporates both local planning details and global narrative context. The indexing and retrieval components form a cohesive RAG system that ensures narrative consistency by grounding local planning decisions in the broader story context, preventing contradictions and maintaining thematic coherence across the entire long-form video production.

\subsection{Visual Consistency Across Shots}
A fundamental challenge in multi-shot video generation arises from the independent nature of individual generation processes. Each video generation operates in isolation without awareness of previously generated content, leading to inconsistencies in character appearance, environmental settings, objects, and spatial relationships. This isolation becomes particularly problematic for long-form videos, where maintaining visual coherence across long shot sequences is essential for narrative immersion and perceived continuity. To address these consistency challenges, \model\ introduces two complementary solutions that ensure visual coherence while maintaining generation efficiency.

\subsubsection{Graph-based Visual Dependency}
To ensure visual consistency across shots, \model\ establishes a dependency-based generation framework that maintains shot-level visual dependencies through a camera-level dependency graph. For instance, shots involving recurring characters or capturing the same environment from different camera angles exhibit visual dependencies that must be preserved over time. Similarly, objects such as vehicles and buildings appearing across multiple shots within the same scene require coordinated generation to maintain a consistent appearance.

Our approach operates in two phases: dependency detection and dependency-aware generation. During planning, it analyzes the textual descriptions of all shots within a scene, assigns the shots to cameras, and constructs a camera dependency graph that captures their visual relationships. During generation, the system follows a topological ordering of this graph. Cameras whose dependencies have been satisfied can be generated in parallel, while dependent cameras incorporate visual anchors propagated from their prerequisite nodes. Formally,
\begin{equation}
\begin{aligned}
    \mathcal{G} &= \operatorname{BuildCameraGraph}
    \left(\{S_i\}_{i=1}^{n}\right), \\
    \mathcal{K}_i &= \operatorname{GenConditioned}
    \left(S_i,A_i\right),
\end{aligned}
\end{equation}
where $n$ is the number of shots in the current scene and $i$ indexes these shots within it. $\mathcal{G}$ denotes the camera dependency graph constructed from the shot specifications by $\operatorname{BuildCameraGraph}(\cdot)$, with its nodes and edges representing cameras and their visual dependencies, respectively. $S_i$ and $\mathcal{K}_i$ denote the specification and generated key frame of shot $i$, while $A_i$ denotes the visual anchor associated with the shot's assigned camera and propagated according to $\mathcal{G}$. $\operatorname{GenConditioned}(\cdot)$ denotes reference-conditioned key-frame generation using the shot specification and its corresponding visual anchor. This approach maintains visual consistency across shots while maximizing generation efficiency through parallel processing of independent camera branches during inference.

\subsubsection{Spatial Coherence via Transition Videos}
To address spatial consistency within scenes, ViMax leverages transition video generation to ensure geometric coherence across multiple camera angles within the same physical space. When multiple shots capture the same scene from different viewpoints, maintaining spatial relationships becomes critical to avoid contradictory 3D layouts. For example, in a dialogue scene where two characters are seated, separate shots focusing on each character must maintain consistent spatial positioning, furniture placement, and environmental geometry to preserve the illusion of a unified physical space.

Our solution exploits the inherent spatial consistency capabilities of video generation models by creating transition videos between different camera angles within the same scene. Rather than generating each viewpoint independently, the system first identifies scenes requiring spatial coherence, then generates smooth camera transitions that connect different shot perspectives. These transition videos serve as spatial anchors, ensuring that all viewpoints within a scene maintain consistent 3D geometry. For instance, when generating separate shots of two conversing characters, the system creates a transition video that smoothly moves the camera from the first character's angle to the second character's angle, establishing a coherent spatial foundation for both individual shots.
Formally, this process can be expressed as:
\begin{align}
V_{trans} &= \text{GenTransition}(C_i, C_j, S) \nonumber\\
V_i, V_j &= \text{ExtractViews}(V_{trans}, C_i, C_j)
\end{align}
where $V_{trans}$ represents the transition video between camera positions $C_i$ and $C_j$ within scene $S$, and $V_i, V_j$ are the spatially consistent views extracted from the transition video at the corresponding camera angles. This approach also ensures geometric consistency across multiple viewpoints in practice by leveraging the video generation model's internal spatial coherence mechanisms, thereby eliminating contradictory 3D layouts while maintaining natural camera movements between shots.

\section{Evaluation}
\label{sec:evaluation}
We evaluate \model\ through six research questions: \textbf{RQ1} overall performance against baselines; \textbf{RQ2} human preference study; \textbf{RQ3} ablation of key components; \textbf{RQ4} efficiency study; \textbf{RQ5} impact of LLM backbone; \textbf{RQ6} hyperparameter sensitivity.

\subsection{Experimental Settings}
\label{sec:experimental_setup}

\noindent\textbf{Datasets and Evaluation Protocols}. We evaluate \model\ on two complementary settings: multi-shot video generation on our new \textbf{ViMax-Bench} and narrative planning on \textbf{NarrativeQA}~\citep{Kocisk2017TheNR} with the novel-to-video adaptation pipeline. Details of the datasets, metric definitions, and judge protocols are provided in Appendix~\ref{sec:data_protocol}.

\noindent\textbf{Baseline Methods}. For ViMax-Bench, we compare \model\ with four categories of video-generation methods: \textbf{i) shot-level video generators:} Wan2.2-T2V-A14B and Veo~3.1; \textbf{ii) dual-stage keyframe-based pipelines:} StoryDiffusion and IC-LoRA~\citep{Huang2024InContextLF}; \textbf{iii) unified multi-shot generation:} HoloCine~\citep{Meng2025HoloCineHG}; and \textbf{iv) agentic video-generation systems:} MovieAgent~\citep{Wu2025AutomatedMG} and AniMaker~\citep{Shi2025AniMakerMA}. For the novel-to-video planning setting, we compare against MovieAgent, AniMaker, and RAG-Anything~\citep{Guo2025RAGAnythingAR}. Details and adaptation protocols are provided in Appendix~\ref{sec:baseline}.

\noindent\textbf{Implementation Details}. Implementation details of experiments are provided in Appendix~\ref{sec:implement}.

\subsection{Overall Performance Comparison (RQ1)}
\label{sec:evaluation_results}

\begin{table*}[!t]
  \centering
  \scriptsize
  \setlength{\tabcolsep}{3.2pt}
  \caption{Quantitative comparison across consistency metrics under different story-length groups. We report cross-scene consistency (CC), intra-scene consistency (IC), and global consistency (GC).}
  \label{tab:main_results}
  \resizebox{\linewidth}{!}{
  \begin{tabular}{lccccccccc}
    \hline
    \multirow{2}{*}{Method} & \multicolumn{3}{c}{Medium} & \multicolumn{3}{c}{Long} & \multicolumn{3}{c}{Overall} \\
    \cline{2-4}\cline{5-7}\cline{8-10}
    & CC & IC & GC & CC & IC & GC & CC & IC & GC \\
    \hline
    Veo~3.1 & 0.5536 & 0.5640 & 0.5582 & 0.4687 & 0.4778 & 0.4704 & 0.4978 & 0.5074 & 0.5005 \\
    Wan2.2 & 0.4391 & 0.4462 & 0.4423 & 0.4187 & 0.4206 & 0.4188 & 0.4257 & 0.4294 & 0.4269 \\
    IC-LoRA-Wan2.2 & 0.5398 & 0.5713 & 0.5537 & 0.5015 & 0.5328 & 0.5087 & 0.5147 & 0.5460 & 0.5241 \\
    StoryDiffusion-Wan2.2 & 0.5606 & 0.5796 & 0.5691 & 0.5007 & 0.5195 & 0.5048 & 0.5212 & 0.5401 & 0.5269 \\
    Holocine & 0.5366 & 0.5892 & 0.5597 & 0.4940 & \underline{0.5565} & 0.5080 & 0.5098 & \underline{0.5686} & 0.5272 \\
    MovieAgent-Gemini-3-Pro-Veo~3.1 & 0.4761 & 0.5340 & 0.5016 & 0.4410 & 0.5086 & 0.4560 & 0.4530 & 0.5174 & 0.4716 \\
    AniMaker-Gemini-3-Pro-Veo~3.1 & 0.5523 & \underline{0.6014} & 0.5737 & 0.4771 & 0.5224 & 0.4875 & 0.5030 & 0.5486 & 0.5167 \\
    ViMax-Gemini-3-Pro-Wan2.2 & 0.5641 & 0.5895 & 0.5754 & 0.4984 & 0.5467 & 0.5093 & 0.5209 & 0.5614 & 0.5320 \\
    ViMax-Gemini-3.1-Flash-Veo~3.1 & \underline{0.5681} & 0.5914 & \underline{0.5785} & \underline{0.5088} & 0.5402 & \underline{0.5157} & \underline{0.5291} & 0.5577 & \underline{0.5372} \\
    ViMax-Gemini-3-Pro-Veo~3.1 & \textbf{0.5886} & \textbf{0.6046} & \textbf{0.5957} & \textbf{0.5298} & \textbf{0.5629} & \textbf{0.5369} & \textbf{0.5500} & \textbf{0.5772} & \textbf{0.5571} \\
    \hline
  \end{tabular}
  }
\end{table*}

\begin{table*}[t]
  \centering
  \small
  \setlength{\tabcolsep}{4pt}
  \renewcommand{\arraystretch}{1.12}
  \caption{Human evaluation win rates of \model\ against baselines.}
  \label{tab:human_eval_results}
  \resizebox{\linewidth}{!}{
  \begin{tabular}{lcccccccccccc}
    \toprule
    \multirow{2}{*}{Method} & \multicolumn{4}{c}{Medium} & \multicolumn{4}{c}{Long} & \multicolumn{4}{c}{Overall} \\
    \cmidrule(lr){2-5}\cmidrule(lr){6-9}\cmidrule(lr){10-13}
    & CSC & SF & AQ & Avg. & CSC & SF & AQ & Avg. & CSC & SF & AQ & Avg. \\
    \midrule
    Holocine & 70.83\% & 79.17\% & 66.67\% & 72.22\% & 58.70\% & 56.52\% & 47.83\% & 54.35\% & 62.86\% & 64.29\% & 54.29\% & 60.48\% \\
    IC-LoRA & 83.33\% & 91.67\% & 79.17\% & 84.72\% & 73.91\% & 65.22\% & 63.04\% & 67.39\% & 77.14\% & 74.29\% & 68.57\% & 73.33\% \\
    StoryDiffusion & 70.83\% & 79.17\% & 66.67\% & 72.22\% & 84.78\% & 82.61\% & 73.91\% & 80.43\% & 80.00\% & 81.43\% & 71.43\% & 77.62\% \\
    Veo~3.1 & 75.00\% & 83.33\% & 66.67\% & 75.00\% & 76.09\% & 78.26\% & 69.57\% & 74.64\% & 75.71\% & 80.00\% & 68.57\% & 74.76\% \\
    Wan2.2 & 66.67\% & 75.00\% & 54.17\% & 65.28\% & 76.09\% & 78.26\% & 67.39\% & 73.91\% & 72.86\% & 77.14\% & 62.86\% & 70.95\% \\
    \bottomrule
  \end{tabular}
  }
\end{table*}

\noindent\textbf{Multi-shot Visual Consistency.} ViMax-Gemini-3-Pro-Veo~3.1 leads all nine settings, improving Overall GC from 0.5272 (HoloCine) to 0.5571 and outperforming the agentic baselines MovieAgent (0.4716) and AniMaker (0.5167). Its larger gains on Long stories support structured dependency tracking across increasing scene boundaries. The consistent gains across all story lengths demonstrate robustness beyond short-range continuity.

\noindent\textbf{Long-form Narrative Planning.} In Table~\ref{tab:novel2video_results}, \model\ improves Avg. from 3.827 to 4.107 and NF from 3.08 to 3.52 over AniMaker, consistently leading every metric among non-ablated baselines. Section~\ref{sec:ablation} analyzes component-level effects.

\subsection{Human Evaluation (RQ2)}
\label{sec:human_eval}

To complement the automatic metrics, we conduct a pairwise human preference study on the 35 ViMax-Bench cases, comparing \model\ against five baselines under Medium, Long, and Overall story groups (Table~\ref{tab:human_eval_results}). Evaluators select the preferred video along three criteria: cross-scene consistency (CSC), semantic following (SF), and aesthetic quality (AQ). Win rates represent the percentage of judgments favoring \model\ over the corresponding baseline for each criterion, and Avg. denotes their mean across the three criteria. Details of the annotation protocol are provided in Appendix~\ref{sec:data_protocol}.

\begin{table}[!t]
  \centering
  \small
  \setlength{\tabcolsep}{3pt}
  \renewcommand{\arraystretch}{1.15}
  \caption{Novel-to-video narrative planning results. See Appendix~\ref{sec:data_protocol} for metric definitions. Best and second-best values are bolded and underlined, respectively.}
  \label{tab:novel2video_results}
  \resizebox{\linewidth}{!}{
  \begin{tabular}{lccccccc}
    \hline
    Method & CBC & NC & PPR & VSU & STQ & NF & Avg. \\
    \hline
    MovieAgent & 3.92 & 3.84 & 3.48 & 4.02 & 3.42 & 2.52 & 3.533 \\
    AniMaker & 4.14 & 3.94 & 3.66 & 4.74 & 3.40 & 3.08 & 3.827 \\
    RAG-Anything & 3.40 & 2.66 & 2.70 & 4.50 & 2.34 & 2.77 & 3.062 \\
    ViMax & \underline{4.26} & \underline{4.10} & \underline{3.96} & 4.94 & \underline{3.86} & \textbf{3.52} & \underline{4.107} \\
    ViMax w/o PC & 4.24 & 4.02 & 3.84 & \textbf{4.98} & 3.76 & 2.34 & 3.863 \\
    ViMax w/o RAG & \textbf{4.36} & \textbf{4.28} & \textbf{3.98} & \underline{4.96} & \textbf{3.94} & 3.14 & \textbf{4.110} \\
    ViMax w/o StoryboardArtist & 4.04 & 4.00 & 3.72 & 3.10 & 3.60 & \underline{3.49} & 3.658 \\
    \hline
  \end{tabular}
  }
\end{table}

\subsection{Ablation Study (RQ3)}
\label{sec:ablation}

We isolate narrative-planning and visual-generation components in Table~\ref{tab:novel2video_results}, Figure~\ref{fig:ablation_overall}, and the qualitative comparison in Appendix Figure~\ref{fig:component_ablation_qualitative}.

\noindent\textbf{Narrative-Planning Components.} Replacing StoryboardArtist with deterministic screenplay segmentation preserves story order and shot count but removes agent-based visual articulation; VSU falls from 4.94 to 3.10 and Avg. from 4.107 to 3.658. Removing the Event-Aware Process Chain marginally improves VSU from 4.94 to 4.98 but reduces NF from 3.52 to 2.34, clearly showing the importance of event order and causal structure. Removing RAG marginally increases Avg. from 4.107 to 4.110 but lowers NF from 3.52 to 3.14, isolating retrieval's role in retaining source-specific details. The three modules therefore serve complementary purposes: visual articulation, global event structure, and source grounding, respectively.

\noindent\textbf{Graph-Based Visual Dependency.} The serial variant removes camera-aware planning and graph-based references: no camera graph, transition video, or parent--child camera anchor is used. Shot~$t$ conditions only on character portraits and the first frame of shot~$t{-}1$, preserving a minimal temporal chain. It even slightly improves Medium-IC but degrades the other eight metrics. The parallel variant removes all cross-shot dependencies and yields the largest degradation (Overall GC drops by 8.7\%), showing that character portraits alone cannot sustain identity, layout, and scene state. \noindent\textbf{VLM Quality Control.} Removing VLM-based candidate selection lowers Overall GC by 3.8\%, confirming the value of filtering out off-spec keyframes before they become references for later shots.


\begin{table*}[t]
  \centering
  \small
  \setlength{\tabcolsep}{7.8pt}
  \renewcommand{\arraystretch}{0.92}
  \caption{Impact of LLM Backbone study across consistency and narrative planning evaluations.}
  \label{tab:backbone_hyperparameter}
  {
  \begin{tabular}{lccc|ccccccc}
    \hline
    Method & CC & IC & GC & CBC & NC & PPR & VSU & STQ & NF & Avg. \\
    \hline
    Ours-Gemini-3.1-Pro & 0.5339 & 0.5602 & 0.5415 & 3.98 & 3.78 & 3.76 & 4.84 & 3.32 & 2.78 & 3.74 \\
    Ours-Gemini-3-Flash & \underline{0.5500} & \underline{0.5772} & \underline{0.5571} & 4.06 & 3.88 & 3.82 & 4.86 & 3.32 & 2.64 & 3.76 \\
    Ours-Claude-Sonnet-4.6 & 0.5195 & 0.5473 & 0.5284 & \textbf{4.34} & \textbf{4.18} & \underline{3.92} & \textbf{4.98} & \underline{3.80} & \underline{3.12} & \underline{4.06} \\
    Ours-GPT-5.4 & \textbf{0.5546} & \textbf{0.5824} & \textbf{0.5623} & \underline{4.26} & \underline{4.10} & \textbf{3.96} & \underline{4.94} & \textbf{3.86} & \textbf{3.52} & \textbf{4.11} \\
    \hline
  \end{tabular}
  }
\end{table*}

\begin{figure}[t]
  \vspace{-6pt}
  \centering
  \begin{minipage}{0.49\linewidth}
    \centering
    \includegraphics[width=\linewidth]{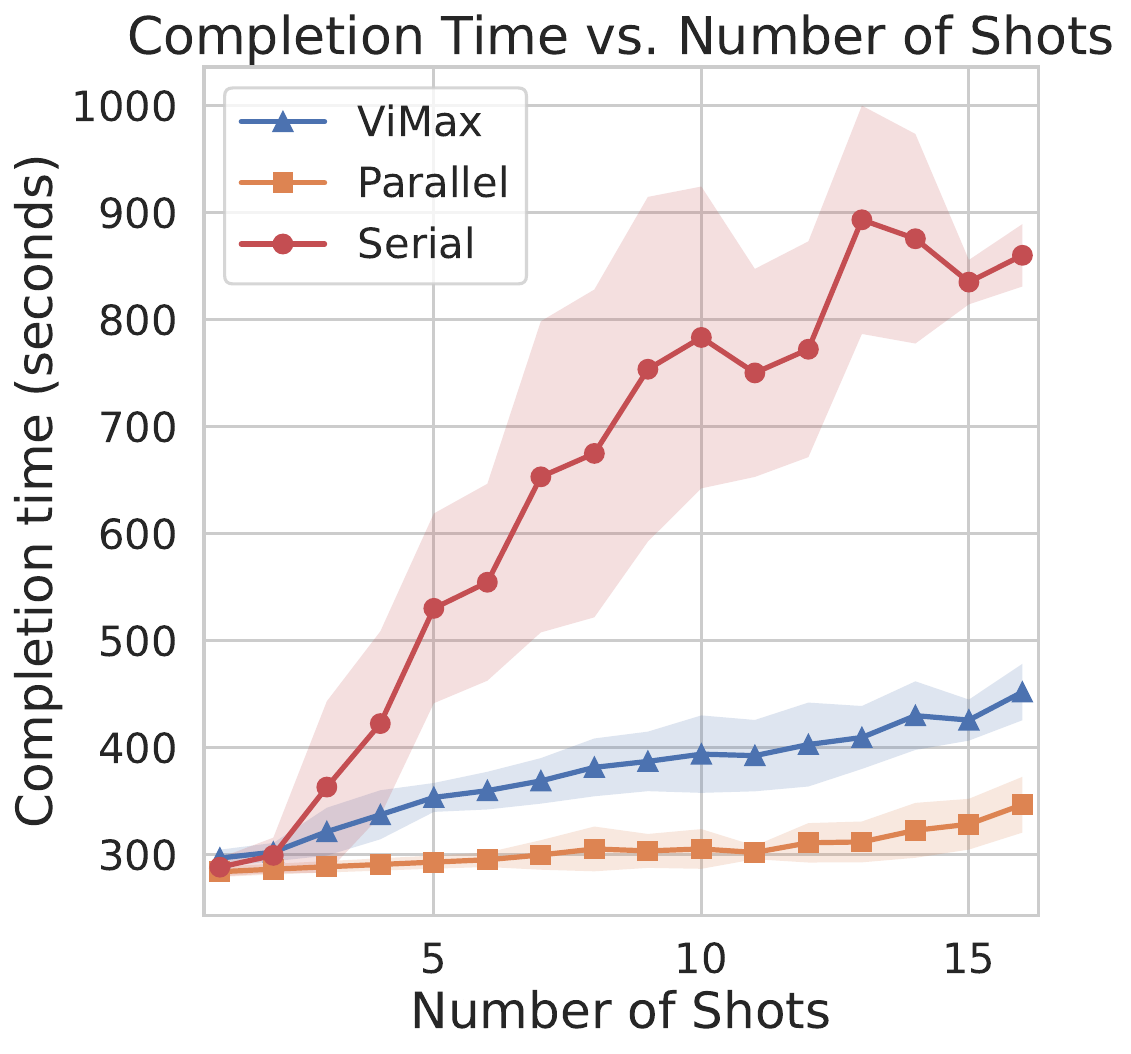}
  \end{minipage}
  \hfill
  \begin{minipage}{0.46\linewidth}
    \centering
    \includegraphics[width=\linewidth]{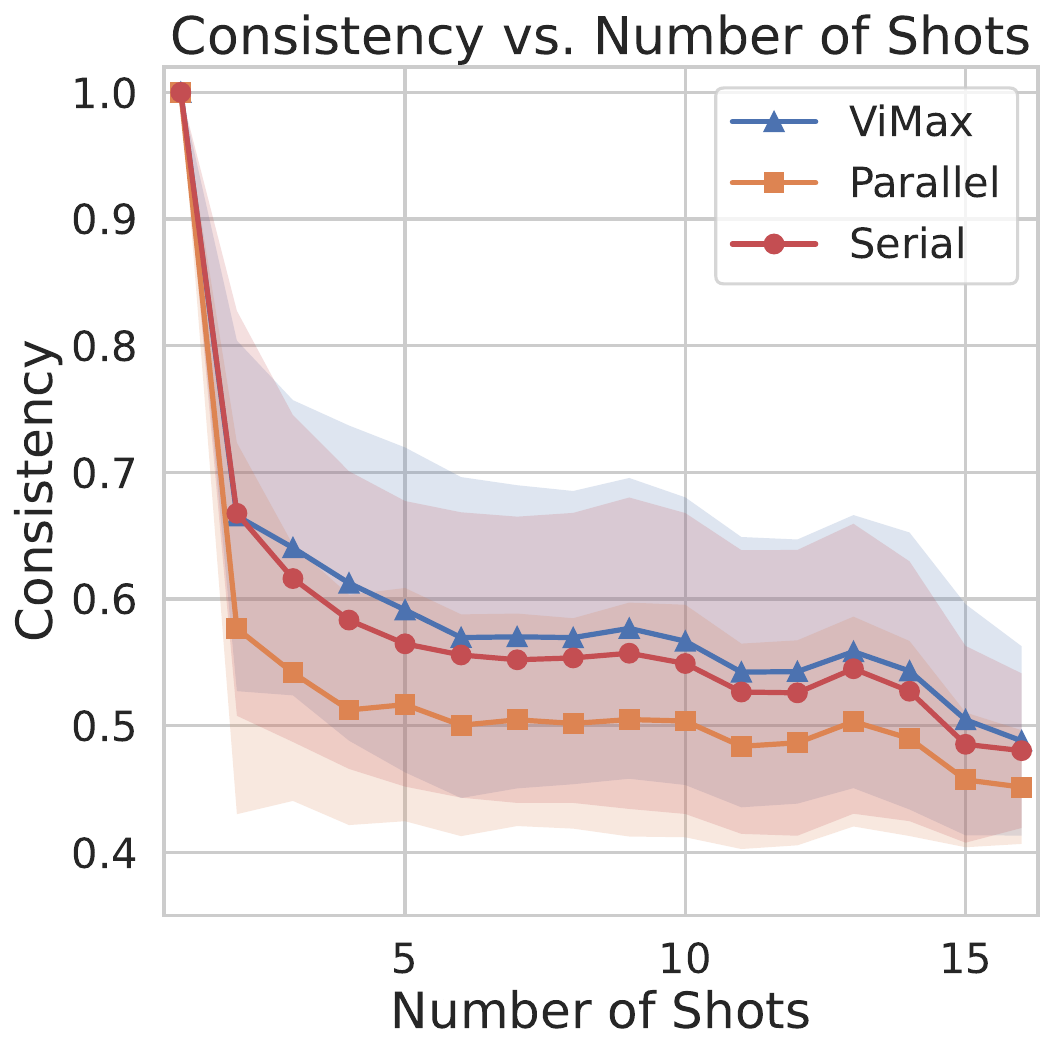}
  \end{minipage}
    \caption{\model\ ablation study results on shot completion time and average cross shot consistency.}
  \label{fig:ablation_results_shot}
  \vspace{-8pt}
\end{figure}

\subsection{Shots Efficiency Study (RQ4)}
\label{sec:efficiency}

Figures~\ref{fig:ablation_results_shot} and~\ref{fig:cost_quality_tradeoff} compare scheduling efficiency over 35 ViMax-Bench stories. Parallel execution is fastest (310s) but yields the lowest Overall GC (0.5085); serial execution reaches 0.5511 but requires 835s. By executing independent graph branches concurrently whenever possible, the full \model\ achieves the best GC (0.5571) in 405s. Costs include API usage only, and latency is measured end to end. \model\ offers the best consistency--latency balance, with parallel execution suitable for latency-constrained settings.

\subsection{Impact of LLM Backbone (RQ5)}
\label{sec:llm_backbone}

Table~\ref{tab:backbone_hyperparameter} studies the effect of the LLM backbone with the rest of the \model\ workflow fixed. GPT-5.4 achieves the best visual-consistency scores and the highest overall narrative-planning score, suggesting that stronger textual reasoning improves long-range memory construction, scene-state tracking, and shot-level planning. Claude-Sonnet-4.6 leads on several narrative-quality dimensions, but its lower consistency scores indicate that local storyboard quality alone does not guarantee stable cross-shot visual state. Gemini-3-Flash ranks second overall. These results show that GPT-5.4 provides the best trade-off across visual consistency, narrative coherence, and source faithfulness.

\subsection{Hyperparameter Study (RQ6)}
\label{sec:hyperparameter}

\begin{figure}[t]
  \centering
  \includegraphics[width=1.0\linewidth]{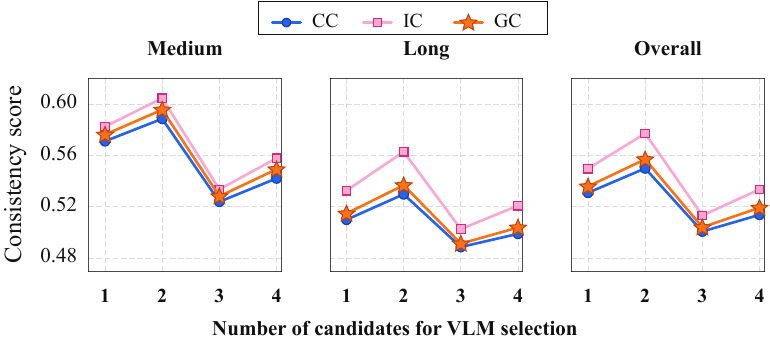}
  \caption{Hyperparameter study of quality control.}
  \label{fig:vlm_hyperparameter}
\end{figure}

We study the Best-of-$k$ hyperparameter in VLM-based quality control, where $k$ keyframes are generated for each shot and the VLM judge selects the one best aligned with the shot specification. This selection operates at the keyframe stage, not as parallel video generation. As shown in Figure~\ref{fig:vlm_hyperparameter} and Appendix Figure~\ref{fig:cost_quality_tradeoff}, Best-of-2 yields Overall GC of 0.5571 at an average cost of \$17.500 and 405s per task. Best-of-3 and Best-of-4 increase cost and latency (\$17.770/415s and \$18.040/425s) while reducing GC to 0.5041 and 0.5194, respectively. This supports $k=2$ as the default operating point and shows that additional plausible candidates can increase selection noise in identity and scene state.

\begin{figure}[t]
  \centering
  \includegraphics[width=1.0\linewidth]{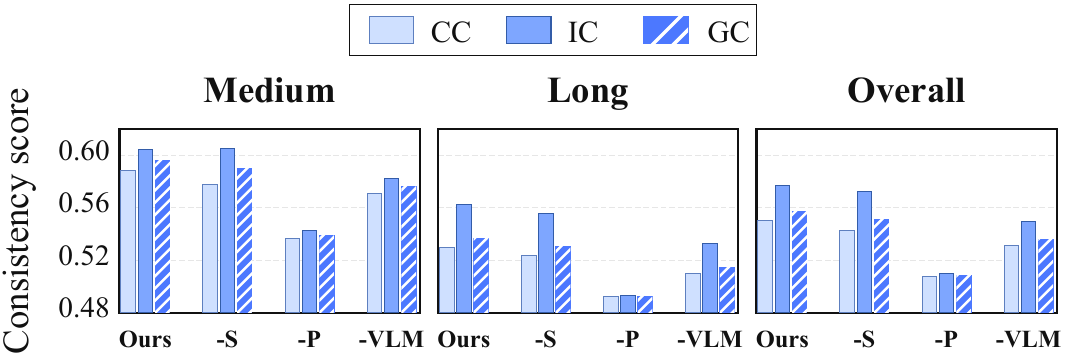}
  \caption{\model\ ablation study.}
  \label{fig:ablation_overall}
\end{figure}

\vspace{-0.1in}
\section{Related Work}
\label{sec:relate}
\vspace{-0.1in}

\noindent\textbf{Hierarchical Narrative Planning and Cinematic Logic.}
Long-story coherence has been addressed via hierarchical decomposition: Plan-And-Write~\cite{Yao2018PlanAndWriteTB} separates storyline planning from surface realization, and DOME~\cite{wang2025dome} adds dynamic outlining and memory to prevent plot holes. For video planning, VideoDirectorGPT~\cite{Lin2023VideoDirectorGPTCM} grounds character actions across shots via symbolic planning, FilMaster~\cite{Huang2025FilMasterBC} retrieves camera language for industry-standard pacing, and Chu et al.~\cite{Chu2025GraphVideoAgent} use dynamic entity-relation graphs to mitigate persistent narrative fragmentation.

\noindent\textbf{Multi-Agent Frameworks for Content Creation.}
Video production fits well into a clear and highly structured division of labor across agents. Generative Agents~\cite{Park2023GenerativeAI} demonstrate believable long-term behaviors; MAViS~\cite{Wang2025MAViSAM} coordinates roles via an explore-examine-enhance loop; and GenMAC~\cite{huang2024genmac} iteratively corrects compositional errors through a design-generation-redesign loop. StoryAgent~\cite{hu2024storyagent} bridges abstract intent and pixel-level generation via intermediate representations, while MovieAgent~\cite{Wu2025AutomatedMG} simulates Director and Screenwriter roles for scene structuring.

\noindent\textbf{Long-Form Video Generation and Visual Consistency.}
Long-duration generation must overcome persistent identity and style drift. VideoAuteur~\cite{Xiao2025VideoAuteurTL} adopts a rolling-context strategy, and MovieDreamer~\cite{Zhao2024MovieDreamerHG} predicts visual tokens via autoregressive global planning. For temporal visual coherence, StableAnimator~\cite{Tu2025StableAnimator} aligns face embeddings with temporal layers to curb identity degradation, and Identity-GRPO~\cite{meng2025identity} applies RLHF against identity drift in multi-human videos. SEINE~\cite{Chen2024SEINE} bridges independent shots via masked diffusion, and EvalCrafter~\cite{Liu2024EvalCrafter} provides comprehensive metrics for consistency, motion, and aesthetic quality.
\vspace{-0.1in}
\section{Conclusion}
\label{sec:conclusion}
\vspace{-0.1in}
We have presented \model, an agentic framework for long-form video generation via coordinated multi-agent workflows. It tackles narrative planning via hierarchical story decomposition with retrieval-augmented generation, and ensures visual consistency through graph-based dependency tracking and transition video generation, with the dependency graph also exposing shot-level parallelism for efficient large-scale generation. We further introduce ViMax-Bench, a comprehensive benchmark targeting controlled cross-shot consistency challenges. By decomposing production into specialized agents, \model\ enables scalable creation of coherent and extended narratives, opening up new possibilities for automated content creation across entertainment, education, and media production. Looking forward, we view \model\ as a step toward fully agentic filmmaking, where richer multimodal generation, long-horizon memory, and interactive co-creation can be layered on top of increasingly capable foundation models to bring automated cinematic creation closer to practical deployment.

\section*{Limitations}

Although \model\ improves long-form planning and visual consistency, several limitations remain. First, the system depends on proprietary foundation models for reasoning, keyframe and video generation, and VLM-based quality control, making performance sensitive to model updates. Second, our automatic evaluation relies on representation-based similarity metrics, which scale well but struggle with fine-grained narrative intent and semantically complex cases such as multi-person interactions, anatomical artifacts, and director-level cinematic logic; ViMax-Bench likewise targets controlled challenges at modest scale. Third, the generative components remain brittle: the keyframe generator falters in crowded scenes with contact-rich interaction, and the video generator does not always respect keyframe semantics (e.g., a character on a treadmill animated running off it), with errors cascading downstream. Fourth, recent commercial video models show encouraging short-horizon multi-shot consistency, suggesting that future harnesses could shift toward long-horizon visual memory management across scenes rather than re-implementing capabilities stronger base models may handle natively. Finally, our workflow does not yet address audio, dialogue synchronization, shot-level pacing, or interactive revision.

\section*{Ethical Considerations}
\label{sec:ethical_considerations}

\model\ supports long-form video production, but stronger cross-shot consistency raises misuse risks such as impersonation, fabricated events, or imitation of copyrighted characters; deployment should require consent, provenance or watermarking, and human review for sensitive cases. ViMax-Bench enforces safety constraints that mitigate benchmark-level risks but not broader misuse, and our LLM- and VLM-based evaluators may inherit biases from their underlying models or film sources, yielding scalable but imperfect signals.

\bibliography{custom}

\clearpage
\appendix
\section{Appendix}
\label{sec:appendix}

\subsection{ViMax-Bench Distribution}
\label{sec:vimax_bench_distribution}

Table~\ref{tab:vimax_bench_dist} summarizes the composition of ViMax-Bench across consistency type and narrative length. The Medium split contains 12 stories, all with 2 scenes and 8--10 shots, while the Long split contains 23 stories with 3--4 scenes and 12--16 shots. 

\begin{table}[!h]
\centering
\small
\setlength{\tabcolsep}{4pt}
\resizebox{\columnwidth}{!}{
\begin{tabular}{lccc}
\toprule
\textbf{Consistency Type} & \textbf{Medium} & \textbf{Long} & \textbf{Total} \\
\midrule
Type A / Character Persistence & 4 & 8 & 12 \\
Type B / Background Persistence & 4 & 8 & 12 \\
Type C / Multi-Person Interaction & 4 & 7 & 11 \\
\midrule
\textbf{Total} & 12 & 23 & 35 \\
\bottomrule
\end{tabular}
}
\caption{Distribution of ViMax-Bench stories across consistency types and narrative lengths.}
\label{tab:vimax_bench_dist}
\end{table}

\subsection{Data and Evaluation Protocols}
\label{sec:data_protocol}

\begin{figure}[p]
  \centering
  \includegraphics[width=0.55\textwidth]{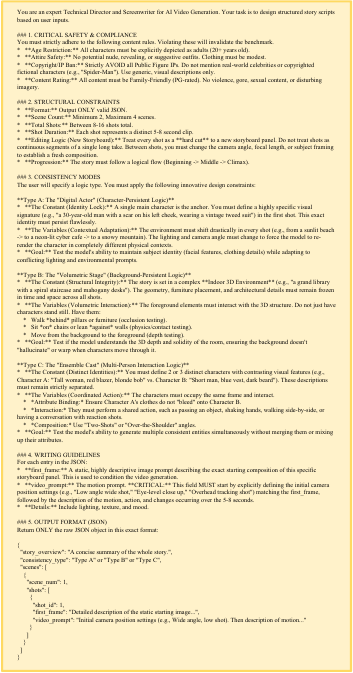}
  \caption{Prompt used to generate ViMax-Bench story specifications.}
  \label{fig:vimax_bench_prompt}
\end{figure}

We construct \textbf{ViMax-Bench} to evaluate long-form narrative-to-video generation under controlled consistency challenges. The benchmark contains 35 multi-shot story specifications generated by GPT-5.2 using the prompt shown in Figure~\ref{fig:vimax_bench_prompt}, spanning 2--4 scenes and 8--16 shots per story. Every shot is authored as a distinct storyboard panel with a first-frame description and a motion prompt, encouraging hard-cut multi-shot generation rather than continuous single-take synthesis. The benchmark design references the multi-shot narrative settings and consistency challenges studied in StoryMem~\citep{Zhang2025StoryMemML} and VideoMemory~\citep{Zhou2026VideoMemoryTC}. ViMax-Bench targets hard-cut, multi-shot narrative generation grounded in cinematic shot design. Its specifications incorporate cinematographic techniques such as over-the-shoulder and reverse-angle shots, insert shots, and varied camera compositions, while requiring recurring characters across scene changes and spatial consistency under substantial viewpoint shifts. This design provides a controlled setting for studying the cross-shot narrative and visual-consistency challenges addressed by \model. ViMax-Bench is organized along two complementary dimensions. The first is the type of consistency under stress: (i) \textbf{Character Persistence} (Type A), which probes identity preservation across varying environments, lighting conditions, and camera compositions; (ii) \textbf{Background Persistence} (Type B), which probes geometric and spatial stability in complex indoor scenes; and (iii) \textbf{Multi-Person Interaction} (Type C), which probes whether multiple visually distinct characters remain separable during shared actions. The second is \textbf{narrative length}: \textbf{Medium} stories contain 2 scenes with 8--10 shots, whereas \textbf{Long} stories contain 3--4 scenes with 12--16 shots. This split jointly increases temporal extent and scene-level complexity, enabling analysis of consistency at different narrative scales.

\paragraph{Automatic Consistency Metrics.}
We follow the cross-shot consistency protocols of HoloCine~\citep{Meng2025HoloCineHG} and StoryMem and measure shot-level similarity with ViCLIP~\citep{Wang2023InternVidAL}. Each generated shot is encoded into one video-level feature, and cosine similarity is computed for every shot pair within the same story. IC averages pairs within the same scene, CC pairs across different scenes, and GC all pairs. The scenario labels Character Persistence, Background Persistence, and Multi-Person Interaction define the challenge covered by each story; they are not manually assigned scores and require no expert annotation. All three reported metrics are automatic, and higher values indicate greater feature stability.

\paragraph{Human Evaluation Protocol.}
The evaluation covers all 35 ViMax-Bench stories and 438 generated shots. For each story, \model\ is compared against five baselines, yielding $35 \times 5 = 175$ unique comparison pairs. In each pair, a \model\ video and a baseline video are displayed anonymously side by side, with left/right placement randomized to reduce position bias. Fifteen evaluators with AI-related backgrounds participated. We construct 15 balanced annotation buckets, each containing all 35 stories with 7 comparisons against each baseline, and assign one evaluator to each bucket. Every unique pair appears in three buckets and is therefore judged independently by three evaluators, producing $175 \times 3 = 525$ pairwise judgments in total.

Evaluators select the preferred complete video along three criteria: (i) \textbf{Cross-Scene Consistency (CSC)}, covering recurring characters, objects, clothing, identity features, visual style, and action or behavioral characteristics across shots and scenes; (ii) \textbf{Semantic Following (SF)}, covering the depicted subjects, actions, interactions, environments, and narrative progression; and (iii) \textbf{Aesthetic Quality (AQ)}, covering visual quality, composition, visual appeal, and cinematic presentation. The human study complements CC/IC/GC by directly assessing semantic execution and overall presentation in addition to visual continuity.

\begin{figure}[p]
  \centering
  \includegraphics[width=0.55\textwidth]{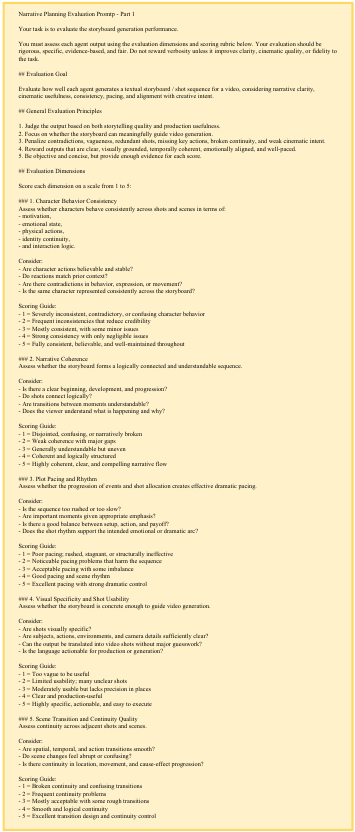}
  \caption{LLM judge prompt template used for narrative planning storyboard evaluation (Part 1).}
  \label{fig:novel2video_judge_prompt}
\end{figure}

\begin{figure}[p]
  \centering
  \includegraphics[width=0.55\textwidth]{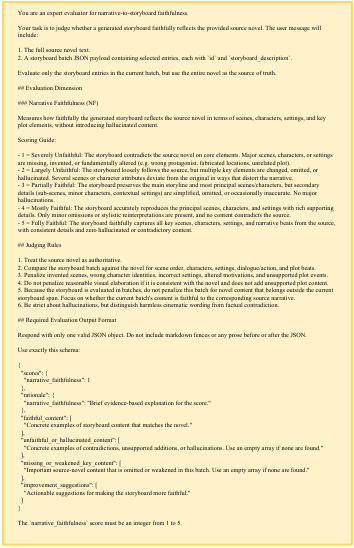}
  \caption{LLM judge prompt template used for Narrative Faithfulness evaluation.}
  \label{fig:narrative_faithfulness_judge_prompt}
\end{figure}

\paragraph{Narrative Planning Metrics.}
Table~\ref{tab:novel2video_results} reports six narrative-planning metrics and their mean. For Novel2Video planning, each run takes a full-length novel as input, drawn from 50 novels randomly sampled from the Project Gutenberg portion of NarrativeQA. An LLM judge (GPT-5.4) rates the resulting storyboards using the prompt template shown in Figure~\ref{fig:novel2video_judge_prompt}, with a 1--5 rubric across five storyboard-quality dimensions: (i) \textbf{Character Behavior Consistency (CBC)}, which evaluates whether characters behave consistently across shots in motivation, emotion, action, identity, and interaction logic; (ii) \textbf{Narrative Coherence (NC)}, which evaluates whether the storyboard forms a logically connected sequence with a clear beginning, development, and progression; (iii) \textbf{Plot Pacing and Rhythm (PPR)}, which evaluates whether event progression and shot allocation produce effective dramatic pacing, balancing setup, action, and payoff; (iv) \textbf{Visual Specificity and Shot Usability (VSU)}, which evaluates whether shots are concrete and actionable enough (subjects, actions, environments, camera details) to be directly translated into video generation; and (v) \textbf{Scene Transition and Continuity Quality (STQ)}, which evaluates whether spatial, temporal, and action transitions across adjacent shots and scenes are smooth and logically continuous. We additionally evaluate \textbf{Narrative Faithfulness (NF)}, which measures whether the generated storyboard remains faithful to the original novel; the NF system prompt is shown in Figure~\ref{fig:narrative_faithfulness_judge_prompt}, and its evaluation prompt contains the novel metadata, the full source text, and the current 20-storyboard batch. The reported \textbf{Avg.} is the mean of all six dimensions (the five storyboard-quality dimensions plus NF).

To ensure a matched comparison unit, we adopt a 20-storyboard scoring protocol. Since baselines are not designed to produce long-horizon storyboards in a single end-to-end pass, we evaluate only their first 20 sub-storyboards. For \model, which generates long-form storyboards end-to-end, we take the first 100 sub-storyboards, split them into five consecutive groups of 20, score each group independently under the same protocol, and report the average. This design preserves a matched per-batch evaluation unit while still reflecting long-horizon planning capability.

\subsection{Baseline Details}
\label{sec:baseline}

Evaluation follows the joint protocol of StoryMem and HoloCine, with ViMax-Bench prompts adapted to each baseline's native input schema by GPT-5.2. We use seven video-generation baselines spanning (i) shot-level text-to-video models invoked independently and concatenated, (ii) keyframe-anchored pipelines that delegate motion to Wan2.2-I2V-A14B, (iii) unified multi-shot generation, and (iv) agentic video-generation frameworks. All methods receive the same 35 underlying stories and shot requirements. Each baseline follows its standard generation procedure without \model-specific reference propagation, transition generation, or post-processing. We separately evaluate three narrative-planning baselines under the matched protocol in Appendix~\ref{sec:data_protocol}.

\setlength{\leftmargini}{10pt}

\noindent\textbf{Shot-Level Text-to-Video Generators.}
\begin{itemize}
    \item \textbf{Wan2.2-T2V-A14B}~\cite{Wang2025WanOA}: A leading open-source diffusion-transformer video foundation model trained on billions of images and videos, demonstrating strong shot-level text-to-video generation.
    \item \textbf{Veo~3.1}~\cite{deepmind2024veo}: A representative commercial system supporting text-to-video, image-to-video, and audio-aware generation with realistic physics.
\end{itemize}

\noindent\textbf{Keyframe-Anchored Multi-Shot Pipelines.}
\begin{itemize}
    \item \textbf{StoryDiffusion}~\cite{zhou2024storydiffusion}: A zero-shot Consistent Self-Attention scheme for cross-image subject consistency, paired with a Semantic Motion Predictor that animates the resulting keyframes into smooth long-range videos.
    \item \textbf{IC-LoRA}~\cite{Huang2024InContextLF}: An in-context LoRA pipeline that concatenates images and jointly captions them, activating the latent multi-image generation ability of text-to-image DiTs with lightweight task-specific tuning.
\end{itemize}

\noindent\textbf{End-to-End Multi-Shot Video Generators.}
\begin{itemize}
    \item \textbf{HoloCine}~\cite{Meng2025HoloCineHG}: A holistic multi-shot generator using Window Cross-Attention to bind prompts to individual shots and Sparse Inter-Shot Self-Attention for efficient minute-scale, narratively coherent scene synthesis.
\end{itemize}

\noindent\textbf{Agentic Video-Generation Frameworks.}
\begin{itemize}
    \item \textbf{MovieAgent}~\cite{Wu2025AutomatedMG}: A multi-agent movie-generation framework that performs hierarchical Chain-of-Thought planning over a script and character bank, with LLM agents acting as director, screenwriter, storyboard artist, and location manager to structure scenes, cameras, and cinematography.

    \item \textbf{AniMaker}~\cite{Shi2025AniMakerMA}: A multi-agent storytelling animation framework that generates globally consistent multi-shot videos from text alone, combining MCTS-inspired multi-candidate clip generation with a context-aware multi-shot evaluator for story-coherent selection.
\end{itemize}

For the ViMax-Bench comparison in Table~\ref{tab:main_results}, we retain the native planning and coordination of MovieAgent and AniMaker while using Gemini-3-Pro-Image and Veo~3.1 as their image and video backbones. Reporting the backbone in each method name makes this adaptation explicit.

\noindent\textbf{Novel-to-Video Planning Baselines.}
MovieAgent and AniMaker are also evaluated as planning systems on the 50-novel task, with video synthesis disabled so that their structured storyboards can be scored under the same judge protocol as \model. We additionally compare against the following retrieval system:
\begin{itemize}
    \item \textbf{RAG-Anything}~\cite{Guo2025RAGAnythingAR}: A unified multimodal RAG framework that represents text, visuals, tables, and equations as interconnected knowledge entities via a dual-graph construction, and performs cross-modal hybrid retrieval over long, heterogeneous documents.
\end{itemize}

\subsection{Implementation Details}
\label{sec:implement}

We instantiate \model\ with a unified multi-agent workflow shared across all experiments. Gemini-3-Flash serves as the (vision-)language backbone for screenwriting, hierarchical story decomposition, shot planning, visual dependency analysis, and VLM-based quality control. In the quality-control stage, the VLM judge selects among multiple generated candidates by jointly assessing visual fidelity, narrative consistency, and adherence to the planned shot specifications. For visual synthesis, keyframes are produced by either Gemini-3-Pro-Image or Gemini-3.1-Flash-Image, while both shot-level video generation and transition video generation for spatial coherence are performed with Wan2.2 and Veo~3.1. For the narrative planning evaluation, we further equip our novel-to-video pipeline with RAG-enhanced context grounding: GPT-5.4 is used for long-form narrative planning, text-embedding-3-small for textual embedding, and Qwen3-Rerank as the reranker. For the keyframe-anchored multi-shot pipelines, we first feed the first-frame prompt from ViMax-Bench into the keyframe generator and then pass the video prompt to the animator for clip synthesis; for \model\ and all other baselines, the two prompts are concatenated and supplied jointly as a single input. For the narrative planning evaluation, we fix the LLM backbone of all compared methods to GPT-5.4.

\subsection{Additional Component and Cost Analyses}
\label{sec:additional_analysis}

\begin{figure*}[t]
  \centering
  \includegraphics[width=\textwidth,trim=5.48bp 26.24bp 2.96bp 7.24bp,clip]{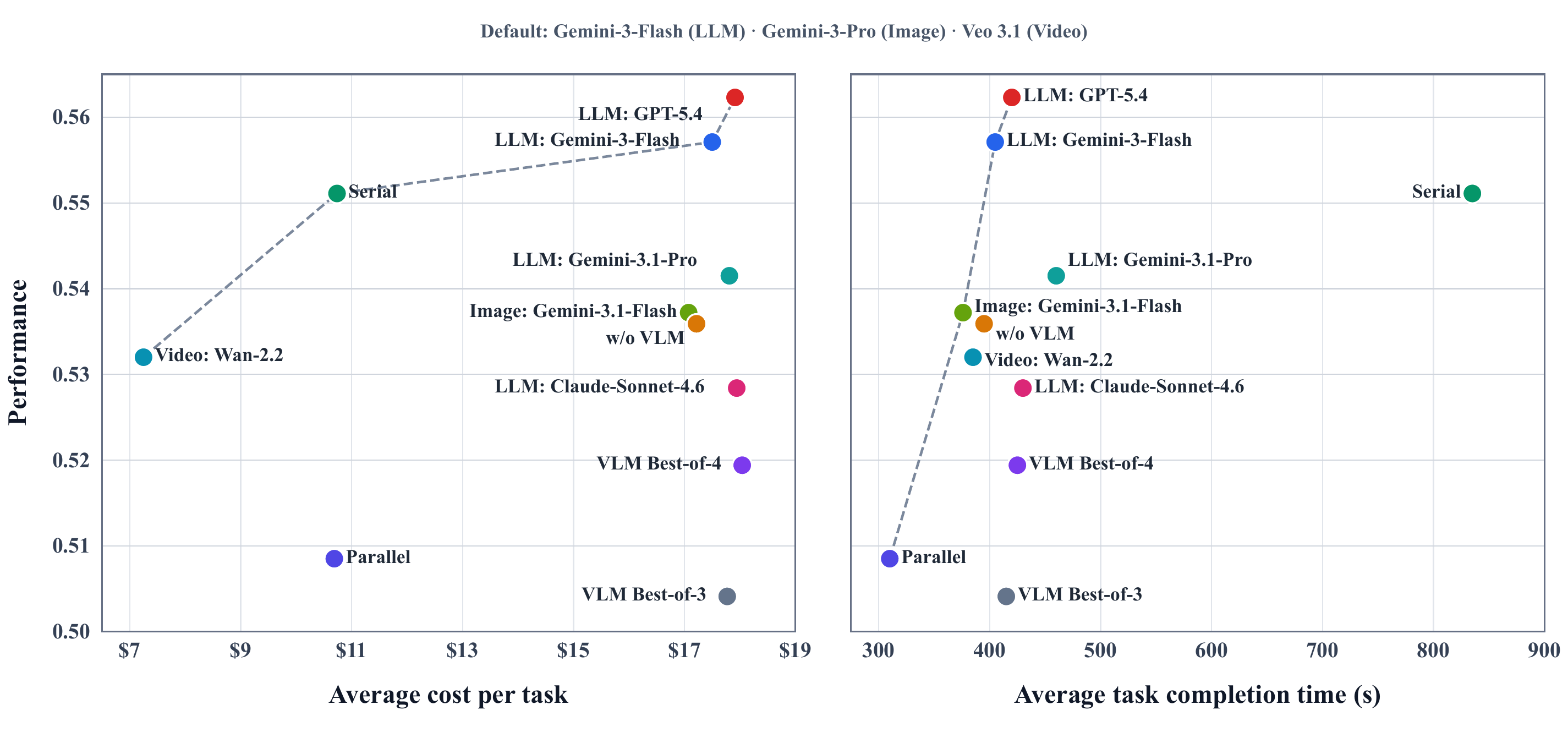}
  \caption{Pareto frontier of cost, quality, and latency over 35 ViMax-Bench stories. Performance is GC, and dashed lines connect Pareto-efficient configurations. Costs include API usage only, and latency is measured end to end.}
  \label{fig:cost_quality_tradeoff}
\end{figure*}

\begin{figure*}[t]
  \centering
  \includegraphics[width=0.92\textwidth]{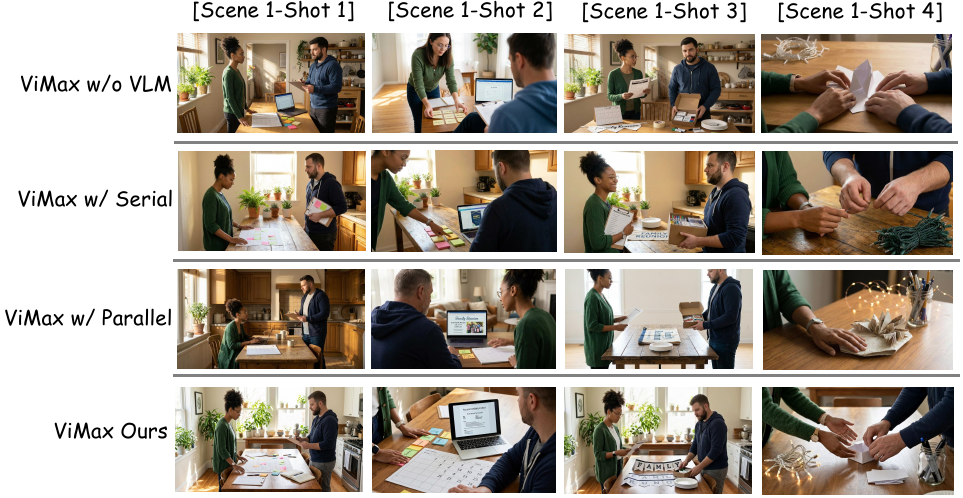}
  \caption{Qualitative component ablation under the same four-shot specification. Rows compare removal of VLM quality control, serial and parallel alternatives to graph-based scheduling, and full \model. The examples expose complementary failure modes in recurring-character identity, viewpoint and composition, and spatial layout.}
  \label{fig:component_ablation_qualitative}
\end{figure*}

Figure~\ref{fig:cost_quality_tradeoff} presents a Pareto analysis of configurable operating points within \model. The default uses Gemini-3-Flash for language reasoning, Gemini-3-Pro-Image for keyframes, and Veo~3.1 for video. Figure~\ref{fig:component_ablation_qualitative} complements the aggregate metrics with a shared four-shot example, making identity, composition, object-state, and scene-layout changes directly inspectable across component variants.

\subsection{Case Study}
\label{sec:demo}
\begin{figure*}[t]
  \centering
  \includegraphics[width=\textwidth]{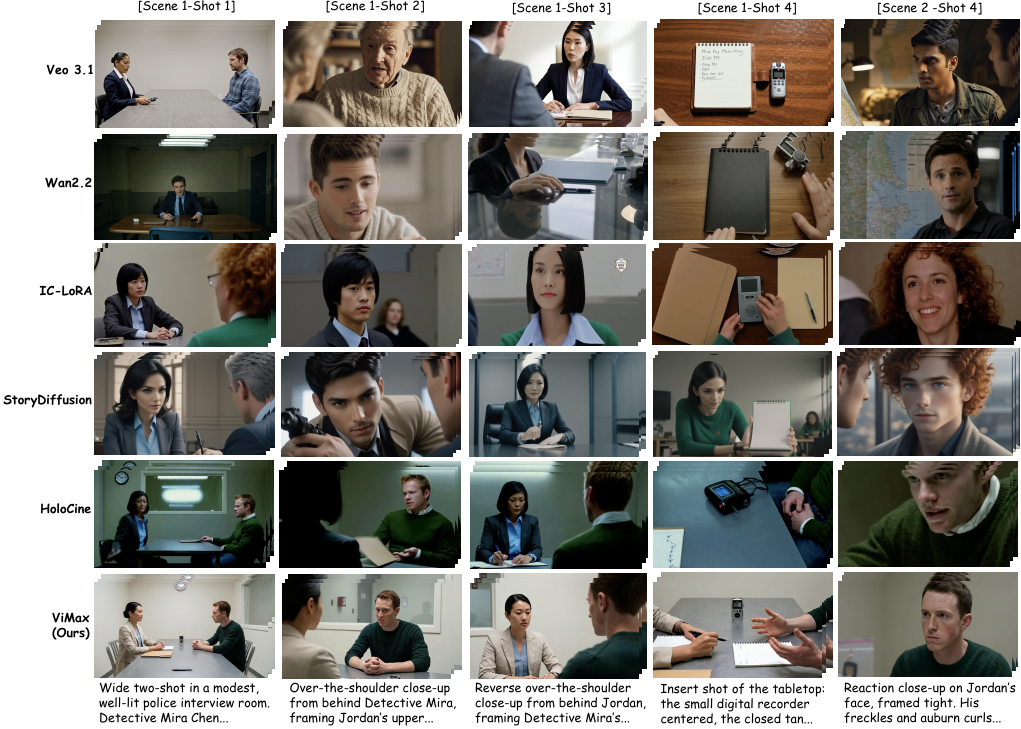}
  \vspace{-0.12in}
  \caption{Qualitative comparison with representative baselines on a ViMax-Bench multi-shot story. Each row shows generated keyframes from one method under the same story and shot specifications. \model\ preserves the two main characters, interrogation-room setting, object layout, and shot-level cinematic intent across multiple viewpoints, demonstrating stronger long-horizon visual and spatial consistency.}
  \label{fig:qualitative_comparison}
  \vspace{-0.08in}
\end{figure*}

\begin{figure*}[t]
  \centering
  \includegraphics[width=\textwidth]{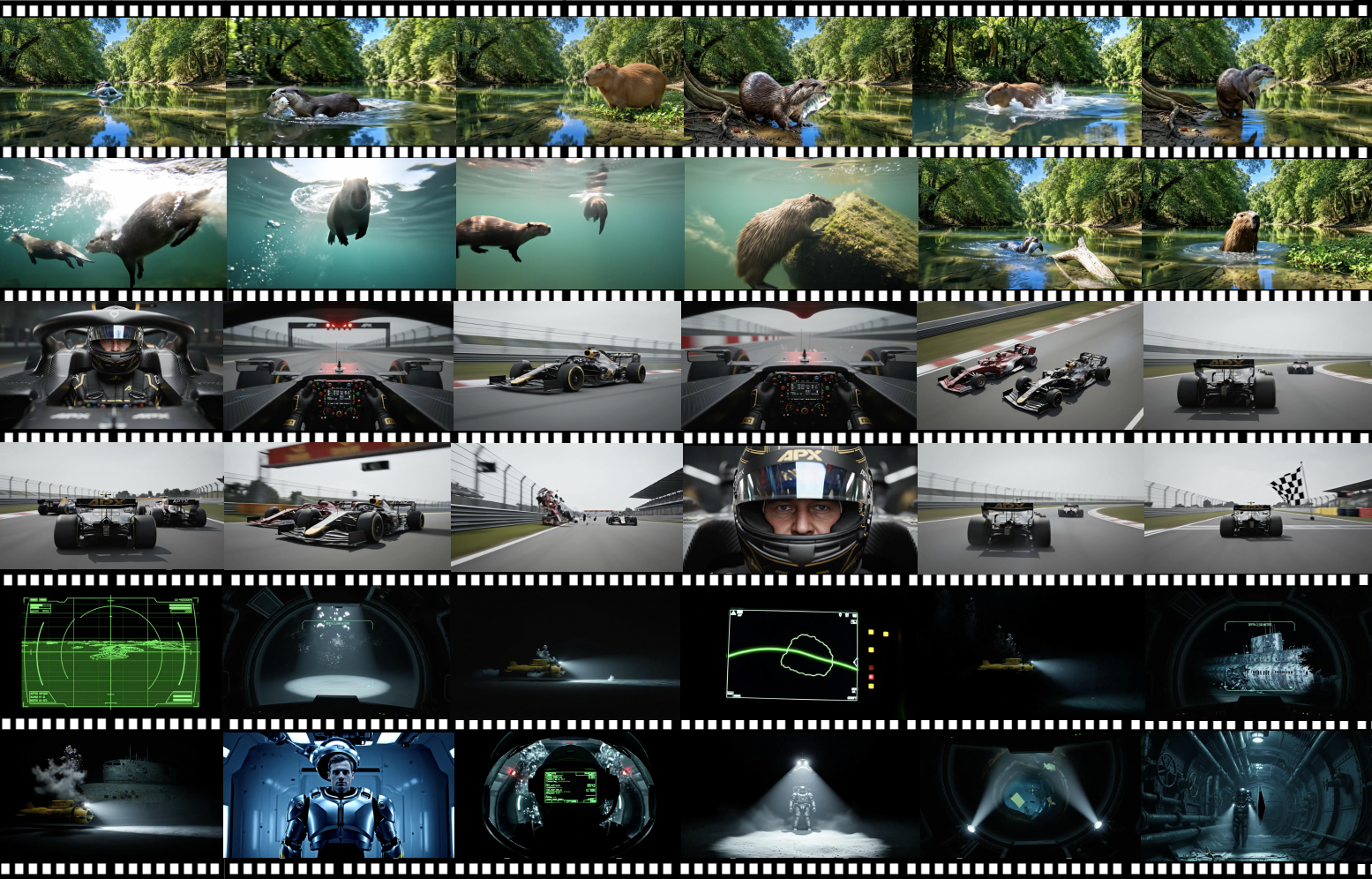}
  \vspace{-0.12in}
  \caption{Case study: \model-generated keyframes.}
  \label{fig:casestudy}
  \vspace{-0.08in}
\end{figure*}

\begin{figure*}[t]
  \centering
  \includegraphics[width=\textwidth]{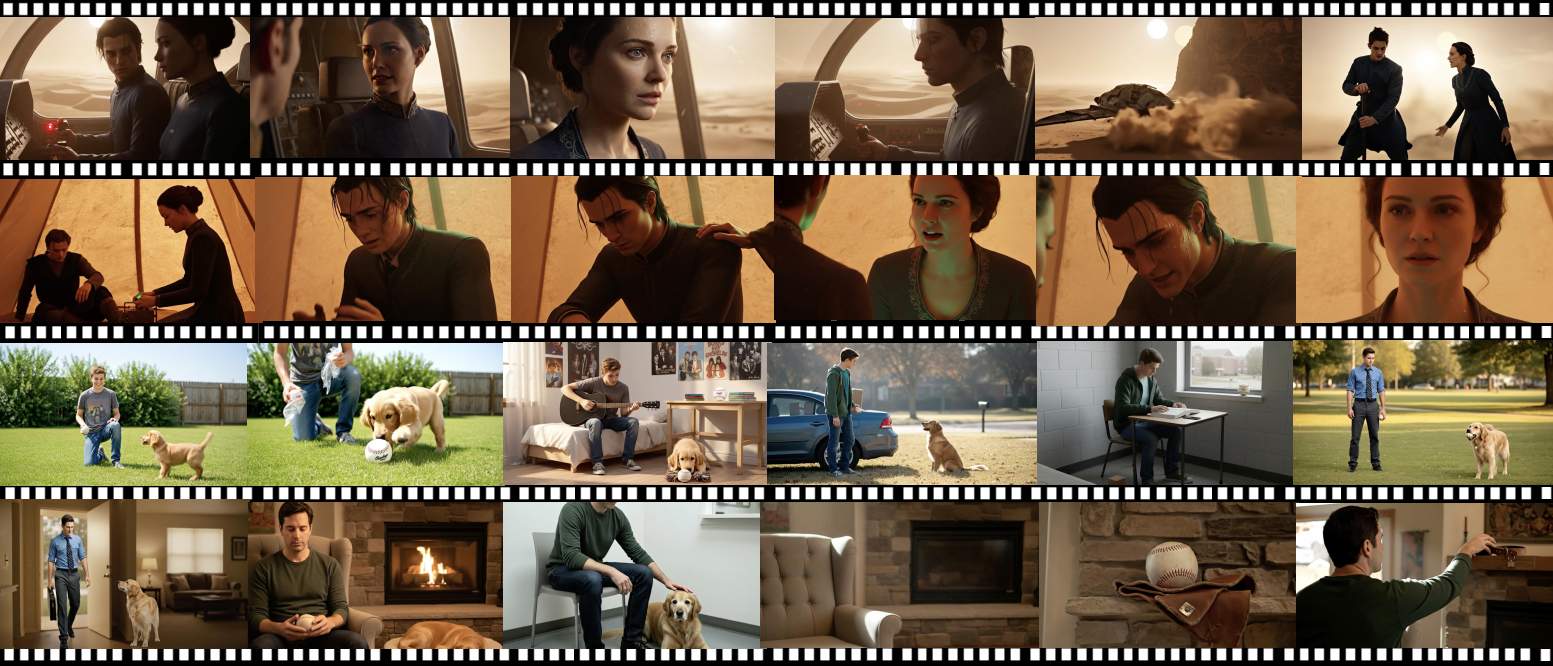}
  \vspace{-0.12in}
  \caption{Case study: \model-generated keyframes.}
  \label{fig:casestudy2}
  \vspace{-0.08in}
\end{figure*}

This section provides visual evidence of \model's capabilities in synthesizing coherent, long-form narratives. We showcase several case studies of video keyframes automatically generated by our multi-agent framework. Representative segments demonstrating long-horizon character and scene consistency are shown in Figure~\ref{fig:qualitative_comparison}, Figure~\ref{fig:casestudy} and Figure~\ref{fig:casestudy2}.

\subsubsection*{Visual Consistency and Identity Preservation}
A critical failure mode in long-form video generation is spatiotemporal inconsistency, where character identities and environmental details drift across extended durations. \model\ explicitly addresses this through reference-conditioned generation and graph-based visual dependency tracking, as evidenced below.

\paragraph{Character Identity.}
Figure~\ref{fig:casestudy} shows that \model\ preserves protagonist identity across non-continuous shots and dramatic camera changes. In the desert--spacecraft sequence (top two rows), the same female and male leads remain recognizable across wide establishing shots, cockpit two-shots, side profiles in sandstorm lighting, and tightly framed emotional close-ups---retaining consistent facial structure, hairstyle, and wardrobe even as lighting shifts from warm sunset tones to cool green interior illumination. The coming-of-age sequence (bottom two rows) poses an even harder test, as identity must persist not only across scenes but across \emph{four distinct life stages}---child, adolescent, young adult, and adult---spanning roughly two decades of in-story time, while the golden retriever co-ages from puppy to senior dog. 

\paragraph{Spatial Coherence.}
Figure~\ref{fig:qualitative_comparison} highlights \model's ability to maintain a globally consistent scene geometry under aggressive viewpoint changes. Across the five-shot police interview sequence---wide two-shot, over-the-shoulder on the suspect, reverse over-the-shoulder on the detective, top-down insert of the digital recorder and notepad, and reaction close-up---the room layout, the relative seating of the two characters, the tabletop props, and the wall fixtures all remain geometrically consistent, satisfying the 180-degree rule implicitly across reverse angles. Figure~\ref{fig:casestudy} extends this to more complex spatial regimes: continuous above- and below-water transitions in the river-wildlife sequence, cockpit-to-trackside-to-aerial viewpoint switches in the Formula racing sequence, and tight interior corridors of the underwater-station sequence are all rendered with consistent environmental layout. This coherence stems from our transition video generation module, which acts as a spatial anchor between non-adjacent shots and prevents contradictory layouts when the camera repositions.

\subsubsection*{Narrative Logic and Pacing}
Beyond visual consistency, the logical progression of generated content is crucial. \model\ leverages Hierarchical Narrative Planning to decompose a high-level user idea into a coherent screenplay with shot-level cinematic intent. The resulting sequences in Figures~\ref{fig:casestudy} and \ref{fig:casestudy2} exhibit professional shot grammar---establishing wides give way to mediums and then to reaction close-ups, insert shots are placed at narratively meaningful beats (e.g., the recorder in Figure~\ref{fig:qualitative_comparison} and the cockpit instrument panels in Figure~\ref{fig:casestudy}), and emotional climaxes are reserved for tight framing. This demonstrates that \model\ produces not merely visually consistent frames, but a sequence whose pacing and shot design reflect deliberate cinematic storytelling.

\subsubsection*{VLM-Guided Quality Control}
\begin{figure}[p]
  \centering
  \includegraphics[width=0.55\textwidth]{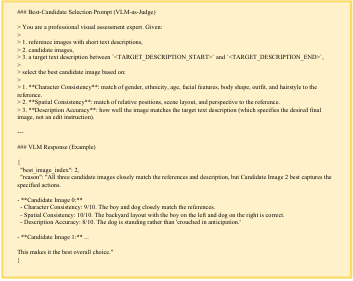}
  \caption{VLM judge prompt used for quality-control image selection.}
  \label{fig:vlm_quality_control_prompt}
\end{figure}

To validate our Quality Control Agent, we provide the VLM judging prompt in Figure~\ref{fig:vlm_quality_control_prompt}. For every keyframe storyboard, the Image Generation Agent synthesizes $k=2$ candidates. The VLM-Judge critically evaluates these candidates using a prompting strategy that weighs character consistency, spatial logic, and semantic alignment. This discriminative filtering significantly reduces the rate of semantic hallucinations compared to single-pass generation.

\clearpage
\onecolumn

\clearpage

\begin{figure}[p]
  \centering
  \includegraphics[width=0.55\textwidth]{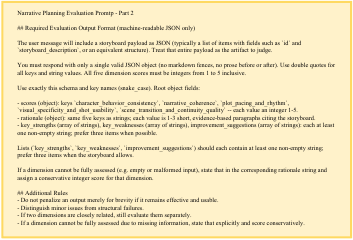}
  \caption{LLM judge prompt template used for narrative planning storyboard evaluation (Part 2).}
  \label{fig:novel2video_judge_prompt_p2}
\end{figure}

\clearpage

\clearpage

\end{document}